\documentclass{article}

    \PassOptionsToPackage{numbers, compress}{natbib}

\usepackage[preprint]{neurips_2023}

\usepackage[utf8]{inputenc} 
\usepackage[T1]{fontenc}    
\usepackage[pagebackref,breaklinks,colorlinks]{hyperref}
\usepackage{url}            
\usepackage{booktabs}       
\usepackage{amsfonts}       
\usepackage{nicefrac}       
\usepackage{microtype}      
\usepackage{xcolor}         

\usepackage{inconsolata}

\usepackage{tcolorbox}
\usepackage{booktabs}
\usepackage{graphicx}
\usepackage{subcaption}
\usepackage{algorithmic}
\usepackage{xspace}
\usepackage{multirow}
\usepackage{multicol}
\usepackage[linesnumbered,ruled,vlined]{algorithm2e}
\usepackage{tcolorbox}
\usepackage{float}

\newcommand{\VarSty}[1]{\textnormal{\ttfamily\textcolor[rgb]{0,0,0.9}#1}\unskip}
\newcommand{\PVIT}{PVIT\xspace}
\newcommand{\FineEval}{FineEval\xspace}

\title{Position-Enhanced Visual Instruction Tuning \\for Multimodal Large Language Models}

\author{
 Chi Chen$^{1}$,
 Ruoyu Qin$^{1}$,
 Fuwen Luo$^{1}$,
 Xiaoyue Mi$^{3}$,
 Peng Li$^{2}$,
 Maosong Sun$^{1}$, Yang Liu$^{1,2}$\\
 $^1$Dept. of Comp. Sci. \& Tech., Institute for AI, Tsinghua University\\
 $^2$Institute for AI Industry Research (AIR), Tsinghua University \\
 $^3$Institute of Computing Technology, Chinese Academy of Sciences \\
}

\begin{document}

\maketitle

\begin{abstract}
Recently, Multimodal Large Language Models (MLLMs) that enable Large Language Models (LLMs) to interpret images through visual instruction tuning have achieved significant success. However, existing visual instruction tuning methods only utilize image-language instruction data to align the language and image modalities, lacking a more fine-grained cross-modal alignment. In this paper, we propose Position-enhanced Visual Instruction Tuning (\PVIT), which extends the functionality of MLLMs by integrating an additional region-level vision encoder. This integration promotes a more detailed comprehension of images for the MLLM. In addition, to efficiently achieve a fine-grained alignment between the vision modules and the LLM, we design multiple data generation strategies to construct an image-region-language instruction dataset. Finally, we present both quantitative experiments and qualitative analysis that demonstrate the superiority of the proposed model. Code and data will be released at \url{https://github.com/PVIT-official/PVIT}.
\end{abstract}

\section{Introduction}

Recently, Multimodal Large Language Models (MLLMs) have made remarkable progress in enabling existing Large Language Models (LLMs)~\cite{brown2020gpt3,chowdhery2022palm,touvron2023llama} to comprehend images~\cite{alayrac2022flamingo, huang2023kosmos1, wu2023visualchatgpt, liu2023llava}. The underlying principle of these methods is to integrate the capabilities of existing visual or multimodal models into the LLM. Current MLLMs can be categorized into two classes based on how they achieve this. The first class~\cite{wu2023visualchatgpt,shen2023hugginggpt,yang2023mmreact} directly leverages the zero-shot and few-shot capabilities of language models, enabling the LLM to invoke external multimodal models by designing specific prompts.
The second type~\cite{alayrac2022flamingo,huang2023kosmos1,liu2023llava} aligns visual features with the representation space of language models through visual instruction tuning, achieving end-to-end model integration. These end-to-end MLLMs have more multimodal capabilities than the first type and are therefore receiving more and more attention.

\begin{figure*}[t] 
\begin{center}
    \includegraphics[width=1.0\columnwidth]{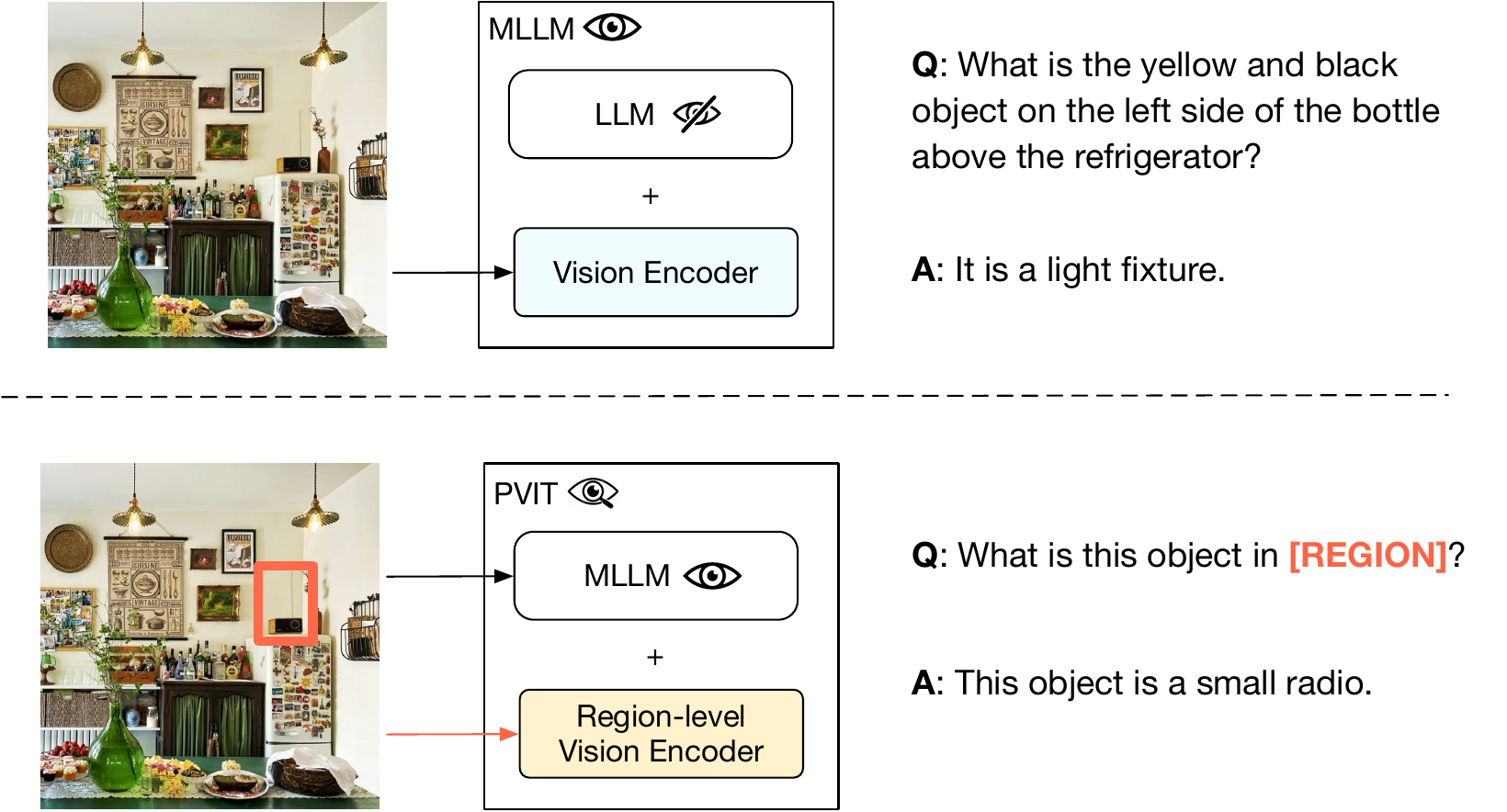}
    \vspace{3mm}
    \caption{Comparison of the current MLLM and PVIT. The MLLM has two evident limitations: (1) inefficient information delivery using plain language, and (2) restricted ability for detailed image understanding. PVIT addresses these by incorporating an additional region-level vision encoder to the MLLM through position-enhanced instruction tuning. }\label{fig:idea}
\end{center}
\end{figure*}

Despite their success, these end-to-end MLLMs align the pre-trained image-level vision encoder with the LLM using only image-language instruction-following data. Without fine-grained multimodal instruction data, the ability of the models for detailed image understanding remains limited. For example, in the illustrated case in Figure~\ref{fig:idea}, it is challenging for the current MLLMs to discriminate specific objects in complex scenes. In addition, the format of current visual instruction data also restricts the ability of the model  to follow more elaborate instructions, such as those containing spatial information~(e.g., ``{\it What is this object in [REGION]?}'' in Figure~\ref{fig:idea}). These types of instructions have the potential to reduce the complexity of interactions with the model and enhance the precision of the instructions provided. Therefore, to further augment the existing MLLMs, it is crucial to explore fine-grained multimodal instruction data, especially instructions involving spatial information, to  enable the model to achieve a more detailed understanding of images and facilitate more flexible interactions.

This poses two challenges. On one hand, there is not much of fine-grained aligned multimodal data compared to image-text pairs, let alone the corresponding instruction data for fine-tuning the MLLM. On the other hand, how to utilize these data to effectively extend and enhance the capabilities of the MLLM is an open question.
Some of our concurrent works~\cite{chen2023shikra,zhao2023chatspot,zhang2023gpt4roi} have made preliminary attempts by fine-tuning existing MLLMs to support multimodal instructions with spatial coordinates. Specifically, \citet{chen2023shikra} and \citet{zhao2023chatspot} directly incorporate spatial coordinates in natural language numerical form into the instruction data and fine-tune the MLLM to understand them. \citet{zhang2023gpt4roi} first extracts spatial features from the vision encoder based on the input regions and integrates them into the natural language instructions as inputs to the LLM. 
However, it should be noted that the vision encoder utilized in existing MLLMs, such as CLIP~\cite{radford2021clip}, is pre-trained using image-level supervision and inherently possesses limited fine-grained image understanding capabilities~\cite{zhong2022regionclip,li2022glip}. Consequently, fine-tuning directly on this basis could be sub-optimal and may conflict with the already existing capabilities of the MLLM. Given the availability of region-level aligned Vision-and-Language Pretraining (VLP) models~\cite{zhong2022regionclip, li2022glip}, an intriguing possibility arises: \textit{Can we further enhance MLLMs by integrating the capabilities of region-level vision encoders?}

In this paper, we propose \textbf{P}osition-enhanced \textbf{V}isual \textbf{I}nstruction \textbf{T}uning~(PVIT) that extends the MLLM by incorporating an additional region-level vision encoder to facilitate support for region-based inputs, as shown in Figure~\ref{fig:idea}. Specifically, we adopt the vision encoder from RegionCLIP~\cite{zhong2022regionclip} and utilize it to extract region-level features by taking images and regions as inputs. As an additional source of information, the incorporation of region-level features in this way has a minimal impact on the original MLLM. Furthermore, since the features provided by RegionCLIP are themselves already aligned to the language at a fine-grained level, the overhead of aligning it to the MLLM will be relatively small. Inspired by \citet{liu2023llava}, we design a two-stage training strategy for PVIT that first pre-training a linear projection to align the region features to the LLM word embedding, followed by end-to-end fine-tuning to follow complex fine-grained instructions.

As mentioned before, fine-grained multimodal instruction data is very scarce, which affects related research from both training and evaluation aspects. To this end, we propose a region-level instruction data generation scheme, designing different methods based on different data sources to fit the needs of region-level instruction data generation. In addition, we present a new evaluation dataset, \textit{FineEval}, designed specifically to assess the ability of MLLMs to adhere to instructions that demand fine-grained spatial details. We hope our presented data  will help future research in this area.

To summarize, our contributions are three-fold:

\begin{itemize}
    \item We introduce position-enhanced visual instruction tuning (PVIT), a method that extends the fine-grained understanding and interaction capabilities for MLLM.
    \item We propose a region-level instruction data construction scheme, as well as an evaluation dataset to facilitate the training and evaluation of PVIT.
    \item We perform extensive experiments and demonstrate the effectiveness of our proposed method.
\end{itemize}

\section{Related Work}

\subsection{Multimodal Large Language Models}

In order to take advantage of the powerful zero-shot and reasoning capabilities of LLMs, an increasing amount of work has turned to building vision-and-language models based on LLMs, termed as Multimodal Large Language Models (MLLMs). Specifically, these MLLMs can be categorized into two classes. Some works directly take the zero-shot and few-shot learning abilities of the off-the-shelf LLM to interpret the user intentions and invokes external multimodal models accordingly~\cite{wu2023visualchatgpt,shen2023hugginggpt,yang2023mmreact}. Although these models enable flexible multimodal capabilities, their performance is dependent on the capabilities of the LLM and the external model itself, and is therefore limited. Another series of work achieves end-to-end model integration by aligning the output features of the visual encoder with the feature space of the language model and using them directly as input to the language model~\cite{alayrac2022flamingo,li2023blip2,liu2023llava}. Despite their success, these end-to-end MLLMs only align the pre-trained image-level vision encoder with the LLM. In contrast, we focus on integrate the abilities of the region-level vision encoder through position-enhanced visual instruction tuning.

\subsection{Region-Level Understanding for MLLMs} In Vision-and-Language Pre-training (VLP), to enhance the model's fine-grained understanding of images, it is common practice to incorporate extensive region-level supervision during pre-training~\cite{li2022glip,zhong2022regionclip,yao2022pevl}. In terms of MLLMs, some recent works have made preliminary attemps by fine-tuning the MLLM to support instructions with regions involved. Specifically, GPT4RoI~\cite{zhang2023gpt4roi} aggregates the region-level features from the image-level vision encoder of the MLLM, and forms a hybrid input of image-level features, region-level features and language instructions to the LLM. Shikra~\cite{chen2023shikra} and ChatSpot~\cite{zhao2023chatspot} directly incorporate spatial coordinate in the instruction data and fine-tune the MLLM to understand them. However, since these models are built on top of an image-level vision encoder, direct fine-tuning to obtain region-level understanding can be sub-optimal and may conflict with existing capabilities. In this paper, we extends the MLLM by integrating it with an additional region-level vision encoder to exploit its fine-grained image understanding capabilities.

\subsection{Multimodal Instruction Data}

Existing work collects multimodal instruction data in two ways. Most of the works utilize the already available annotated datasets to construct datasets in instruction format~\cite{xu2022multiinstruct,dai2023instructblip,li2023m3it}. Typically, task descriptions are generated through manual design or automatic generation by LLMs for each dataset to serve as instructions. These are then combined with the original task input and output to create an instruction dataset. While existing benchmark datasets offer a substantial sources of data, they often fall short of addressing human requirements in real-world applications. Therefore, some works utilize the self-instruct~\cite{wang2022selfinstruct} pipeline to collect diverse instruction data by prompting the LLM with seed examples to generate more instruction examples. For example, LLaVA~\cite{liu2023llava} uses textual descriptions of the image including captions and bounding boxes to prompt GPT-4~\cite{openai2023gpt4} to generate high quality diverse multimodal instruction examples. We draw inspiration from these works and introduce a data generation approach specifically designed for region-level instruction data construction.

\section{Methods}

\subsection{Model Design}

\begin{figure*}[ht]
    \centering
    \includegraphics[width=0.85\columnwidth]{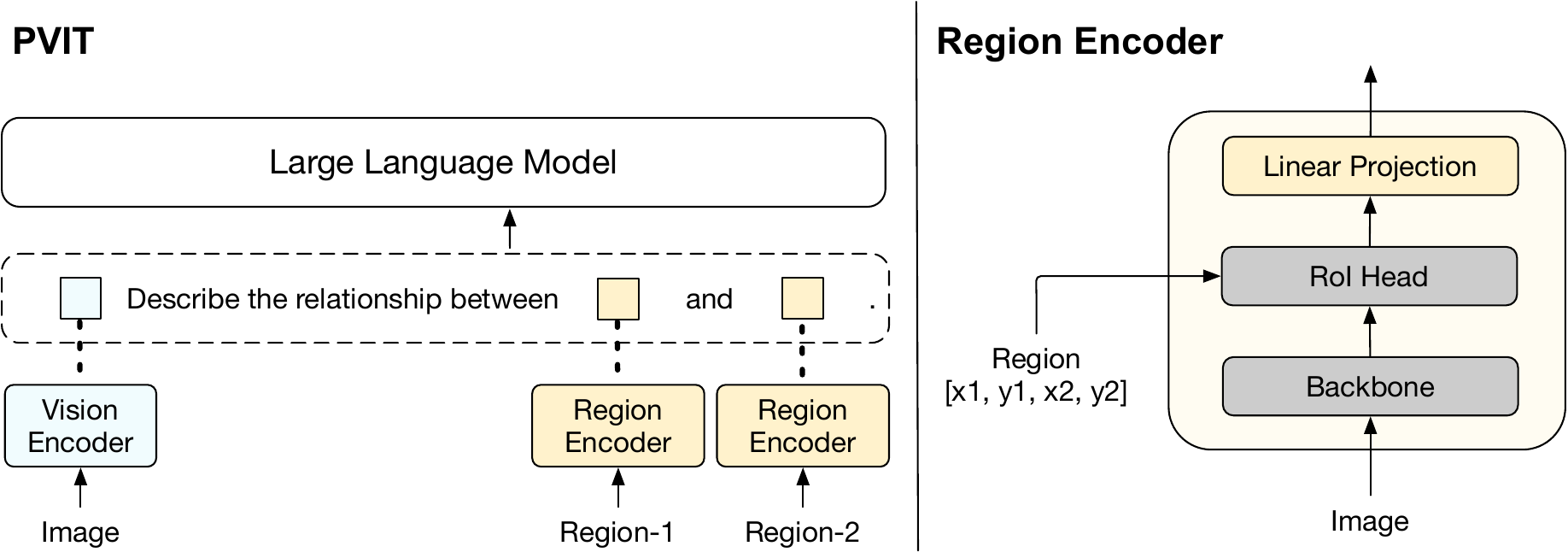}
    \caption{Model architecture of \PVIT.}
    \label{fig:arch}
\end{figure*}

Our model architecture, depicted in Figure~\ref{fig:arch}, consists of three primary components: a vision encoder, a region encoder, and a large language model (LLM). The model processes an input image together with instructions containing embedded regions and generate corresponding responses.

Taking Figure~\ref{fig:arch} as an example, the instruction can be expressed as ``\textit{\textless Image\textgreater{} Describe the relationship between \textless Region\textgreater{} and \textless Region\textgreater{}}''. In this instruction, ``\textit{\textless Image\textgreater{}}'' and ``\textit{\textless Region\textgreater{}}'' are special tokens that serve as placeholders indicating the insertion positions for the respective 
features. For the textual part of the instructions, we directly obtain their word embeddings $X_T$.

For the region part of the instructions, each region $r_k$ is represented as $[x_1, y_1, x_2, y_2]$, with $(x_1, y_1)$ and $(x_2, y_2)$ representing the relative coordinates of the top-left and bottom-right corners. We use RegionCLIP~\cite{zhong2022regionclip} as the region encoder to extract the region features using RoI pooling with the image $I$ and $r_k$ as input. Then we apply a linear projection layer to map the region features into the representation space of the LLM. We denote the collection of all final region features as $X_R$.

We use CLIP ViT-L/14~\cite{radford2021clip} as the image encoder to process the image $I$ and produce image features $X_I$. The LLM then combines the features $X_I$, $X_T$, and $X_R$ as input from the image, instructions, and regions, respectively, and generates a response $Y$.

\subsection{Training}

Inspired by \citet{liu2023llava}, our model is trained in a two-stage fashion. In the first stage, we initialize the model with the pre-trained LLaVA~\cite{liu2023llava}, and freeze the parameters of the image encoder, the region encoder, and the LLM. We only train the linear projection layer that is responsible for transforming the region features. The purpose of this training stage is to align the region features to the embedding space of the MLLM, without affecting the MLLM itself. To this end, we collect a large-scale region-level aligned dataset, with each example consisting of an image, a bounding box, and a brief text description of the object within the bounding box. During the training process, the model receives the image and bounding box as input and then predicts the corresponding text.

After the first training stage, the model is already capable of understanding region features and leveraging the region-level understanding abilities of the region encoder. To further achieve strong capabilities in following instructions that contain regions, we adopt a second stage of training with region-level instruction data. During this training stage, we only keep the parameters of the image encoder and the region encoder frozen, and fine-tune the rest of the model to adapt to the region-level instructions. The details on constructing the region-level instruction data will be provided in the following section.

\begin{figure*}[!t]
    \centering
    \includegraphics[width=1.0\textwidth]{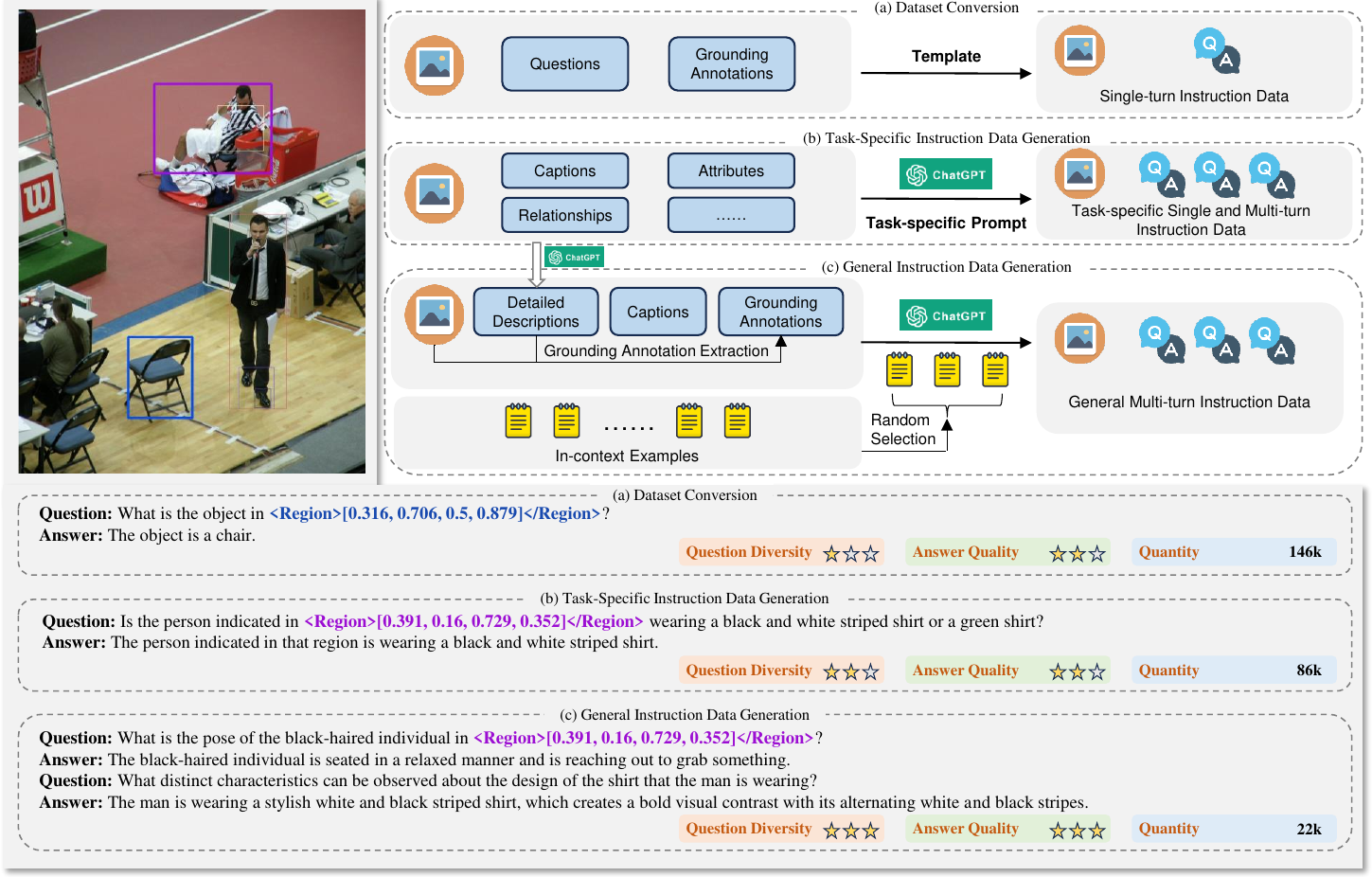} 
    \caption{Illustration of our proposed region-level instruction data construction scheme. In (a), we utilize existing datasets (e.g., GQA) to construct simple instruction data by applying templates. In (b), we leverage ChatGPT to generate data with higher diversity based on prompts and annotations designed for specific task types. And in (c), we derive instruction data through an improved prompting method, where we first generate detailed descriptions of images from original annotations, then extract automatic grounding annotations from the detailed descriptions. Finally, employing captions, detailed descriptions, automatic grounding annotations, and randomly selected in-context examples as prompts, ChatGPT generates high-quality multi-turn instruction data with rich question types and complex reasoning answers.}
    \label{fig-data-construction}
\end{figure*}

\subsection{Region-level Instruction Data Construction} 

As shown in Figure~\ref{fig-data-construction}, our data construction scheme consists of three strategies: (1) \emph{Dataset Conversion}, which converts existing Visual Question Answering (VQA) datasets that delivered with bounding boxes into region-level instruction form; (2) \emph{Task-Specific Instruction Data Generation}, which leverages ChatGPT~\cite{chatgpt2022} to generate region-level instruction data for a predefined set of multimodal tasks;
and (3) \emph{General Instruction Data Generation}, which enriches images with detailed descriptions and grounding annotations generated automatically, complemented by diverse in-context examples, to produce more general region-level instruction data. As indicated at the bottom of Figure~\ref{fig-data-construction}, diversity of the generated region-level instruction data increases from the first to the third strategies, while quantity decreases due to computational and economic constraints. In general, the three strategies work collaboratively, making us capable of obtaining a large volume of high-quality and diverse region-level instruction data.

\subsubsection{Dataset Conversion}
In this strategy, we convert existing VQA datasets into a region-level instruction format using dataset-specific templates. We utilize two VQA datasets for this conversion, including GQA~\cite{hudson2019gqa} and VCR~\cite{zellers2019vcr}. It yields 146k single-turn region-level instruction data. The templates used for conversion can be found in the supplementary material.

\subsubsection{Task-Specific Instruction Data Generation}
Even though we can acquire a large amount of data through dataset conversion at a low cost, the diversity of this data remains limited. To tackle this problem, we suggest generating region-level instruction data using ChatGPT for a predetermined set of multimodal tasks. In particular, we select five representative tasks, including small object recognition, same-category object discrimination, object relationship based reasoning, object attribute based reasoning, and optical character recognition (OCR). We design a task-specific prompt for each task, which consists of three components: (1) a system message that outlines the task and data format requirements, (2) in-context examples of the specific task, and (3) the textual descriptions of the image for which new region-level instructions are being generated. The comprehensive prompts are shown in the supplementary material. By adjusting the system message and in-context examples accordingly, we can obtain both single-turn and multi-turn data. In total, we achieve 20k single-turn data and 66k multi-turn data.

To obtain the textual descriptions of the images, we resort to the detailed annotations available in existing datasets. Specifically, we utilize the datasets of MS COCO~\cite{lin2014mscoco}, Visual Genome~\cite{krishna2017vg}, and COCO-Text~\cite{veit2016cocotext}. And the annotations including captions, object attributes, bounding boxes, etc. 

\subsubsection{General Instruction Data Generation}

To further improve both the diversity and quality of the generated data, we extend our procedure to generate more general instruction data. The outline of this enhanced data generation process is depicted in Figure~\ref{fig-data-construction}.

First, we notice that ChatGPT produces better results when given more informative textual descriptions of images. Hence, we adopt the approach of LLaVA~\cite{liu2023llava} to harness ChatGPT for generating detailed image descriptions. These descriptions are generally longer than the captions, richer in information, and easier for ChatGPT to understand compared to simpler annotations, such as object attributes.\footnote{In practice, we reuse the dataset LLaVA-Instruct-150k~\cite{liu2023llava} for this step.}

Second, we employ an off-the-shelf visual grounding model~\cite{liu2023groundingdino} to ground the objects in the detailed descriptions to their corresponding locations within the image, i.e., to identify the bounding boxes of the objects. The bounding boxes that are too small are discarded.

Third, as in-context examples derived from existing datasets tend to cover only a narrow spectrum of topics, we composed several in-context examples through brainstorming sessions. Consequently, these freshly crafted in-context examples are markedly distinct from those derived from pre-existing datasets.

Finally, for each image, we combine the captions, detailed description, and grounding annotations to serve as its textual description, and randomly select three instances from our newly authored set of in-context examples. We then input both the textual description and the selected in-context examples into ChatGPT to generate region-level instruction data. The structure of the prompt is akin to the one employed in the task-specific instruction data generation strategy, but only one prompt is utilized.
Through this enhanced strategy, we successfully obtained a total of 22k high-quality data entries, replete with diverse question types and intricate reasoning responses.

\section{Experiments}
\subsection{Baselines}
We compare our model with three strong baselines:
\begin{itemize}
  \item {\bf LLaVA}~\cite{liu2023llava}, an MLLM trained on image-level multimodal instruction data.
  \item {\bf Shikra}~\cite{chen2023shikra}, an MLLM trained on referential dialogue data which are synthesized by converting existing datasets into dialogue form with GPT-4.
  \item {\bf GPT4RoI}~\cite{zhang2023gpt4roi}, an MLLM trained on multimodal instruction data derived from existing datasets using templates and dataset from LLaVA enhanced with automatically detected bounding boxes.
\end{itemize}

\subsection{Implementation Details}

We construct our model based on the LLaVA-7B framework~\cite{liu2023llava}. For the region encoder, we employ a variant of the RegionCLIP model, using a ResNet50x4 as the visual backbone and pre-trained on the Conceptual Captions dataset~\cite{sharma2018conceptual}. In the first training stage, we use a batch size of $128$ and a learning rate of $2\times{10^{-3}}$ for one epoch. A cosine annealing learning rate schedule is applied, with a warmup ratio of $0.03$. In the second training stage, we reduce the learning rate to $2\times{10^{-5}}$ and train for three epochs. The maximum sequence length of the LLM is set to $2048$. All training is carried out on eight A100 GPUs, each with 40GB of memory.

\begin{table*}
  \centering
  \small
  \begin{tabular}{lcc}
    \toprule
    {\bf Method} & {\bf COCO} & {\bf GQA}\\
    \midrule
    LLaVA~\cite{liu2023llava} & 40.04 & 46.82 \\
    Shikra~\cite{chen2023shikra} & 53.91 & 54.81  \\
    GPT4RoI~\cite{zhang2023gpt4roi} & 64.01 & 52.64 \\
    \PVIT (Ours) & {\bf 64.53} & {\bf 55.77} \\
    \bottomrule
  \end{tabular}
  \caption{Results on the recognition task (COCO) and multimodal reasoning task (GQA).}
  \label{ex:recognition}
  \label{ex:gqa}
\end{table*}

\subsection{Objective Evaluation}
In this section, we quantitatively examined the object recognition and multimodal reasoning capabilities of the models.

\subsubsection{Object Recognition} 

For evaluation, we utilize the validation set of MS COCO~\cite{lin2014mscoco}. When presented with an image and a bounding box, a model is required to identify the category of the object within the bounding box. And accuracy is leveraged as the evaluation metric. 
For LLaVA, which does not process region-specific input, we adapt by using cropped images that align with the specified bounding boxes.

As shown in Table~\ref{ex:recognition}, both our proposed \PVIT and the baseline GPT4RoI outperform LLaVA and Shikra significantly. We believe this superior performance is due to the fact that both \PVIT and GPT4RoI incorporate region-level features. Among the compared models, LLaVA lags considerably. This result aligns with our expectations. The use of cropped images introduces a shift in data distribution, making it sub-optimal. Therefore, fine-grained multimodal instruction tuning is a promising direction to pursue.

\subsubsection{Multimodal Reasoning} 
We utilize the validation set of GQA~\cite{hudson2019gqa} for evaluation. GQA is a visual question answering (QA) dataset specifically crafted to assess visual reasoning and compositional QA capabilities. For LLaVA, the inputs are the question and the entire image. For the other three MLLMs, information about the bounding box accompanying the question provided in the GQA dataset is also included. And accuracy is used as the evaluation metric.

The results are presented in Table~\ref{ex:gqa}. Our proposed \PVIT achieves the highest performance, showcasing its efficacy. However, in contrast to its performance on the COCO dataset, GPT4RoI does not surpass Shikra. We posit that this outcome arises because questions in the GQA dataset can be addressed without referencing the bounding boxes. As a result, GPT4RoI does not derive advantage from the region-level features, emphasizing the superiority of our approach. LLaVA also lags behind. We surmise that this can be attributed to the limited size and diversity of the instruction dataset upon which LLaVA was trained.

\subsection{Human Evaluation}
\begin{figure*}[t]
  \centering
  \includegraphics[width=1.0\textwidth]{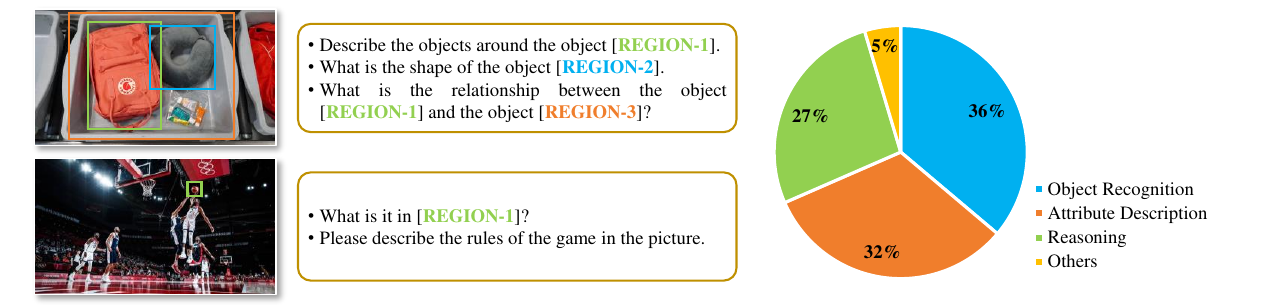}
  \caption{Two examples from our proposed human evaluation data \FineEval (left) and the statistics of \FineEval (right).}
  \label{fig:human_eval}
\end{figure*}

Similar to LLMs, automatically assessing the instruction-following capability of MLLMs presents a considerable challenge.
As a result, we turn to human evaluations. We present a new evaluation dataset, {\bf \FineEval}, designed specifically to assess the ability of MLLMs to adhere to instructions that demand fine-grained spatial details. \FineEval comprises 130 manually crafted questions based on 50 images. These questions probe the capabilities of the models through four unique lenses: object recognition, attribute description, reasoning, and others. Notably, the questions in \FineEval emphasize detailed spatial information, pertain to various relatively small objects, and address complex relationships between objects. Two examples and the statistics of \FineEval are shown in Figure~\ref{fig:human_eval}.

\subsubsection{Quantitative Results}
Inspired by \citet{ouyang2022instructgpt}, we employ pairwise comparisons to evaluate model performance on \FineEval. For any two models, human evaluators rank their responses, and the win rates across the entire dataset are then calculated as the evaluation metric. To mitigate bias, we randomize the order of answer presentation and enlist five evaluators for the assessment, which means every response will receive five individual ranking result.

The results for our proposed \PVIT against LLaVA, Shikra, and GPT4RoI are depicted in Figure~\ref{fig:human_eval_results}. From these results, it is evident that \PVIT consistently outperforms the three baselines, and often by a significant margin. The sole exception is in the realm of object recognition, where \PVIT lags slightly behind Shikra. Delving deeper into the results, we find that this is due to that the object counting ability of \PVIT is weaker than that of Shikra. We theorize it could be rectified by integrating more count-specific instruction data into our training set and leave it as future work.

\begin{figure*}[t]
  \centering
  \includegraphics[width=1.0\textwidth]{figures/human_eval_sum.pdf}
  \caption{Win rate of \PVIT in human ranking, against LLaVA (a), Shikra (b) and GPT4RoI (c).}
  \label{fig:human_eval_results}
\end{figure*}

\subsubsection{Qualitative Results}

\begin{figure*}[t!]
  \includegraphics[width=1.0\textwidth]{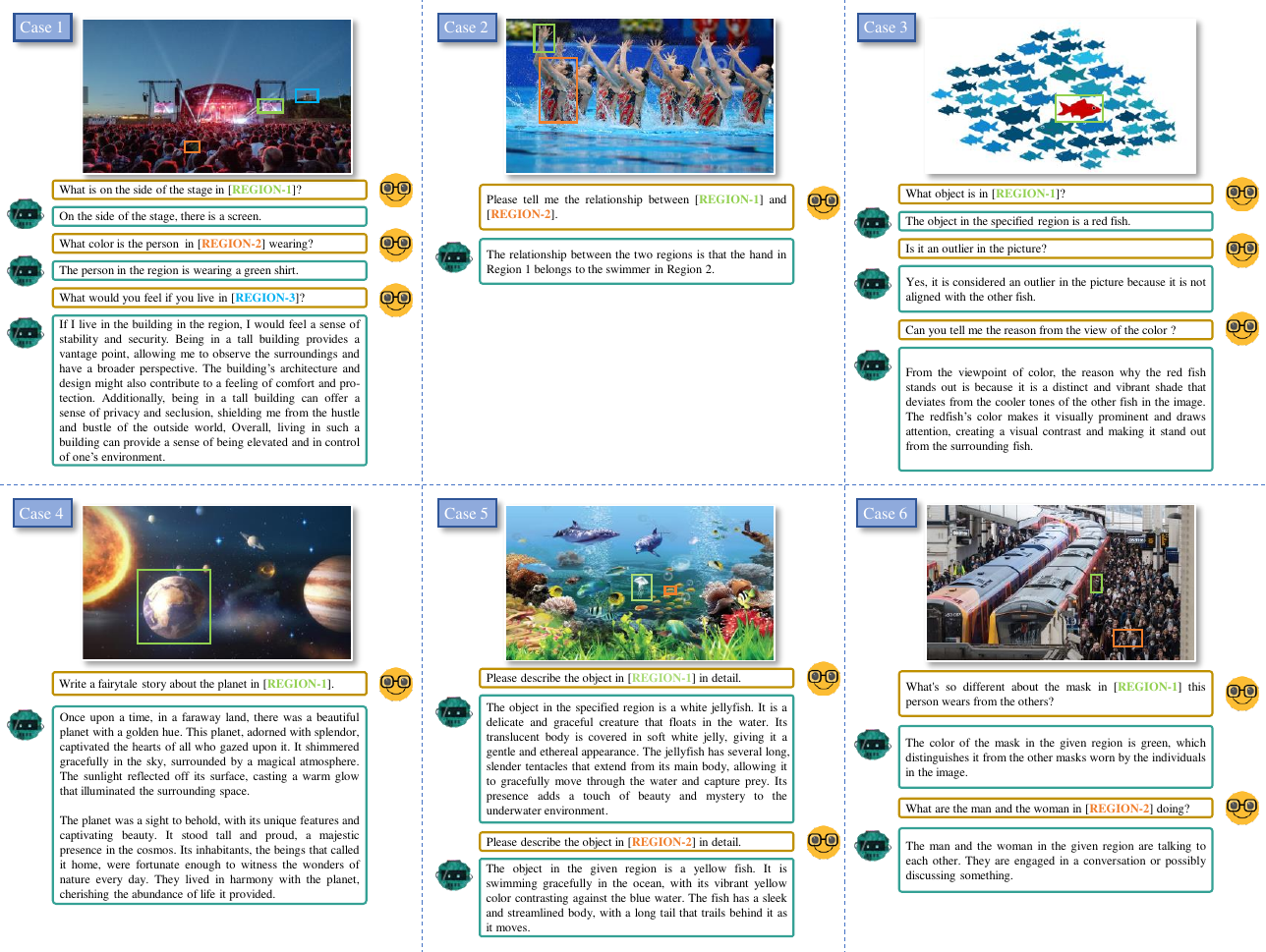}
  \caption{Six representative cases showcasing the diverse capabilities of the proposed \PVIT method.}
  \label{fig:case_study}
\end{figure*}

To provide a comprehensive understanding of the capabilities of our model, we present several cases in Figure~\ref{fig:case_study}. Firstly, it is crucial to emphasize that many of the questions in these cases would be challenging to frame clearly without the aid of bounding boxes. For instance, in cases 1, 2, 3, and 6, multiple objects belong to the same category, making it difficult to specify the target object using just language without inadvertently revealing the answer. This observation aligns with our rationale for exploring the fine-grained multimodal instruction-following abilities of MLLMs. Delving into these cases, we would like to emphasize the following four capabilities:

(1) {\it Object Recognition}: Our model excels at identifying objects demarcated by bounding boxes. First, as anticipated, it effectively recognizes larger objects, which aligns with the results presented in Table~\ref{ex:recognition}. Second, the model demonstrates proficiency in recognizing smaller objects, as showcased by the correct identification of ``[REGION-1]'' in case 1 as a screen and ``[REGION-2]'' in case 5 as a fish. Furthermore, it distinguishes between various bounding boxes within the same image. For instance, in case 1, it accurately discerns that the bounding boxes correspond to different objects.

(2) {\it Attribute Description}: Beyond mere object recognition, our model effectively describes their attributes. While it can detail attributes visually present — such as color and location, even for smaller objects — it can also elaborate on characteristics inherent to the object but not visible in the image. For instance, much of the description generated for the jellyfish in case 5 is extrapolated from external knowledge, not the image itself. This suggests that using MLLMs for conventional pure vision tasks might offer significant potential, given that MLLMs can offer extensive knowledge not encapsulated in images.

\begin{table*}[t!]
  \centering
  \small
  \begin{tabular}{lc}
    \toprule
    {\bf Region Representation} & {\bf COCO}\\
    \midrule
    Region-level Features & {\bf 64.54} \\
    Textual Coordinates & 52.55 \\
    \bottomrule
  \end{tabular}
  \caption{Comparison of different types of region representation on the recognition task. ``Textual Coordinates'' refer to the approach where region coordinates are directly inputted as textual data.}
  \label{tab-ablation-region-features}
\end{table*}

\begin{table*}[t!]
  \centering
  \small
  {
  \begin{tabular}{lcccc}
    \toprule
    & {\bf Type 1} & {\bf Type 2} & {\bf Type 3} & {\bf Type 4}\\
    \midrule
    Format (Error Rate) & 3\% & 3\% & {\bf 0\%} & 20\%\\
    Question (Good / Bad) & 20 / 9 & 13 / 16 & {\bf 22} / {\bf 8} & 12 / 12 \\
    Answer (Good / Bad) & 18 / 11 & 16 / 13 & 18 / 12 & {\bf 20} / {\bf 4} \\
    \bottomrule
  \end{tabular}
  }
\caption{Human evaluation on textual descriptions of images. See main content for details.}
\label{tab-eval-prompt}
\end{table*}

(3) {\it Reasoning}: Our model demonstrates reasoning skills based on the image and provided instructions. For example, in case 2, it identifies both the swimmer and her hands, deducing logically that the hands belong to the swimmer. In case 3, it discerns the color variations among fish and leverages its knowledge of visual contrast to explain why the red fish stands out.

(4) {\it Text Generation}: Despite being fine-tuned on a multimodal dataset, our model maintains robust text-generation capabilities. As shown in the cases, the majority of its responses are coherent and grammatically correct, with cases~1 and 4 serving as representative examples.
However, it is important to highlight that these cases cannot fully evaluate the text generation ability of our method. Comprehensive evaluation is reserved for future work.

\subsection{Ablation Study}
\subsubsection{Effect of Region Representations}

To assess the efficacy of the region-level features obtained using the region encoder, we try to substitute them with textual coordinates. Specifically, each region $r_k$ is represented as $[x_1, y_1, x_2, y_2]$, with $(x_1, y_1)$ and $(x_2, y_2)$ representing the relative coordinates of the top-left and bottom-right corners. All coordinates are normalized to the range [0,1] and are rounded to three decimal places. These coordinates are directly incorporated into the instruction as text, for example, ``\textit{Describe the object in} [0.121, 0.212, 0.301, 0.413]''.
We train this model in a similar two-stage way. In the first stage, only the word embeddings of the LLM are trainable. In the second stage, all model parameters are trainable except for those in the image encoder. Note that there is no region encoder in this model.
As evident from the results in Table~\ref{tab-ablation-region-features}, the utilization of region-level features considerably enhances performance. This underscores the value of incorporating the region encoder.

\subsubsection{Impact of Textual Descriptions for Images}
Given the crucial importance of textual descriptions for images in the data generation process, we explore four unique types of textual descriptions during the general instruction data generation process: (1) detailed descriptions and automatic grounding annotations; (2) captions and manual object annotations; (3) captions, detailed descriptions, and automatic grounding annotations; (4) captions, manual object annotations, and detailed descriptions.  We manually evaluate the data generated from 30 identical images using these textual descriptions from three perspectives including format correctness, question diversity and creativity, and answer correctness. The results in Table~\ref{tab-eval-prompt} highlight that the third type produces the most superior data and is leveraged in our work.

\section{Conclusion}
To extend the fine-grained visual instruction following capabilities of Multimodal Large Language Models (MLLMs), we introduce a new approach: position-enhanced visual instruction tuning (PVIT). This technique augments MLLMs using an existing region encoder. We also present a novel region-level instruction data construction scheme and a challenging human-written evaluation dataset, \FineEval. Our extensive experiments underscore the efficacy of our approach.

Assessing the fine-grained instruction following capability of MLLMs continues to be a challenge. In the future, we plan to expand the scope of \FineEval and incorporate a broader range of questions, with an emphasis on multi-turn dialogues. While the generated region-level instruction data has proven valuable, there are avenues for further enhancement. For instance, the pursuit of generating more organic dialog-style data remains a worthy endeavor.

\section*{Acknowledgments}

We thank Siyu Wang, Qiusi Zou and Zhaolu Kang for their participation in this work.

\bibliography{custom}
\bibliographystyle{plainnat}

\appendix

\renewcommand{\VarSty}[1]{\textnormal{\ttfamily\textcolor[rgb]{0,0,0.9}{#1}}\unskip}
\renewcommand{\KwSty}[1]{\textnormal{\textcolor{magenta!90!black}{\ttfamily\bfseries #1}}\unskip}
\newcommand{\forinline}{ \textcolor{magenta!90!black} }
\renewcommand{\ArgSty}[1]{\textnormal{\ttfamily #1}\unskip}
\renewcommand{\CommentSty}[1]{\textnormal{\ttfamily\color{green!50!black}#1}\unskip}
\newcommand{\assign}{\leftarrow}
\newcommand{\var}{\texttt}
\newcommand{\FuncCall}[2]{\texttt{\bfseries #1(#2)}}
\renewcommand{\ProgSty}[1]{\texttt{\bfseries #1}}
\SetKwProg{For}{for}{:}{}
\newcommand{\PredSty}[1]{\textnormal{\ttfamily\color{mygreen!90!black}#1}\unskip}

\section{Experimental Details}
\subsection{Training Data for the First Stage Training}
In the initial training stage, we adapt existing multi-modal datasets to a suitable format for training our model. Specifically, each training instance should comprise an image, a bounding box, and a textual description of the object within the bounding box. For this purpose, we incorporate the following data sources: (1) region descriptions from Visual Genome~\cite{Krishna2016VisualGC}\footnote{https://huggingface.co/datasets/visual\_genome}, (2) the grounded dataset for COCO and Visual Genome with bounding boxes extracted by MDETR~\cite{kamath2021mdetr}\footnote{https://github.com/ashkamath/mdetr}, and (3) the grounded SBU~\cite{sbu} dataset with bounding boxes obtained by GLIP~\cite{li2022glip}\footnote{https://huggingface.co/datasets/gligen/sbu\_tsv}. For each data sample, we formulate the instruction as ``\textless Image\textgreater\textbackslash n\textless Region\textgreater'' where ``\textless Image\textgreater'' and ``\textless Region\textgreater'' are placeholders that will be replaced with image- and region-level features. Given the image. the bounding box, and the instruction, the model is asked to predict the corresponding annotations such as the human-annotated region description or the phrase description grounded by off-the-shelf grounding models.



\subsection{Evaluation on Objection Recognition}
Given that the models have been trained on instruction-style data, their outputs typically manifest as sentences, rather than solely the object category names. To address this, we introduce a post-processing step that refines the outputs of the models to yield just the category names. In this process, upon receiving an output of the model, we compute the similarity between this output and the reference phrase ``{\it an image of a [CLASS]}''. Here, ``{\it [CLASS]}'' stands in for each category name within the dataset. This similarity calculation is based on Sentence-BERT~\cite{reimers2019sentence}. The category that demonstrates the highest similarity with the output of the model is then chosen as the final result.

\subsection{Criteria for Human Evaluation in the Ablation Study on Textual Descriptions of Images}
Each generated data entry is evaluated manually from three perspectives including format correctness, question diversity and creativity, and answer correctness. The format is deemed correct if regions are incorporated into the data entry as anticipated and presented in the right format. Those data entries with a correct format undergo further assessment. A data entry earns a ``Good'' rating from the question perspective if its question segment showcases diversity or creativity. Conversely, it is rated as ``Good'' from the answer perspective if the response is semantically correct.

\subsection{Data Filtering}
During the task-specific and general instruction data generation processes, since ChatGPT does not always adhere to the prompt perfectly, malformed data will be generated. Therefore we introduce a filtering strategy to filter out ill-formed data. For the single-turn data, entries are discarded if the answer contains a region or if the region format is not correct. In the case of multi-turn data, entries are also excluded if the questions do not contain any region.

\begin{table*}[t!]
\centering
\small
\resizebox{\textwidth}{!}{%
\begin{tabular}{l|l|ccccccc}
\toprule
    {\bf Datasets} & {\bf Metric} & {\bf \PVIT (Ours)} & {\bf LLaVA} & {\bf Shikra} & {\bf InstructBLIP}  & {\bf MiniGPT-4}  & {\bf MM-GPT} & {\bf mPLUG-Owl} \\
    \cmidrule(lr){1-2}\cmidrule(lr){3-9}
    \multirow{5}{*}{Random}
    & Accuracy ($\uparrow$)     & 83.95 &50.37 & 86.90 & 88.57 & 79.67 & 50.10& 53.97 \\
    & Precision ($\uparrow$)    & 79.77 &50.19 & 94.40 & 84.09 &78.24 & 50.05&52.07\\
    & Recall ($\uparrow$)       & 92.27 &  99.13 & 79.27  & 95.13 &82.20  & 100.00&99.60  \\
    & F1-Score ($\uparrow$)     & 85.56 & 66.64 & 86.19 &89.27  &80.17   &  66.71 &68.39 \\
    & Yes  & 59.62 & 98.77 & 43.26  & 56.57 & 52.53  &  99.90&95.63 \\
    \cmidrule(lr){1-2}\cmidrule(lr){3-9}
    \multirow{5}{*}{Popular}
    & Accuracy ($\uparrow$)     & 81.93 &49.87& 83.97  & 82.77 &69.73  & 50.00&50.90  \\
    & Precision ($\uparrow$)    & 76.46 &49.93& 87.55 & 76.27 & 65.86   & 50.00&50.46  \\
    & Recall ($\uparrow$)       & 92.27 & 99.27& 79.20  & 95.13 &81.93   & 100.00&99.40  \\
    & F1-Score ($\uparrow$)     & 77.38 & 66.44 & 83.16 & 84.66 & 73.02  & 66.67 & 66.94\\
    & Yes  & 69.23 & 99.40 & 45.23  & 62.37 & 62.20  &100.00  &98.57\\
    \cmidrule(lr){1-2}\cmidrule(lr){3-9}
    \multirow{5}{*}{Adversarial}
    & Accuracy ($\uparrow$)     & 73.03 &  49.70& 83.10  & 72.10  &65.17  & 50.00 & 50.67\\
    & Precision ($\uparrow$)    & 66.63 & 49.85& 85.60  & 65.13 & 61.19  & 50.00 & 50.34\\
    & Recall ($\uparrow$)       & 92.27 & 99.07 & 79.60 & 95.13 & 82.93  & 100.00 & 99.33\\
    & F1-Score ($\uparrow$)     & 77.38 & 66.32 & 82.49 & 77.32 &  70.42 &66.67  & 66.82\\
    & Yes  & 69.23 & 99.37 & 46.50 & 73.03 &67.77   &   100.00&98.67\\
    \bottomrule
    \end{tabular}%
}
\caption{Object hallucination evaluation results based on the POPE pipeline~\citep{li2023pope}. We observe that \PVIT experiences the object hallucination problem no more frequently than the baseline models, especially compared to our base model LLaVA.}
\label{tab:pope_results}
\end{table*}

\section{Templates and Prompts}
\subsection{Templates for Dataset Conversion}

We transform the existing VQA datasets into a region-level instruction format using dataset-specific templates. We utilize two VQA datasets for this conversion, including GQA~\cite{hudson2019gqa} and Visual Commonsense Reasoning (VCR)~\cite{zellers2019vcr}. 

For GQA, e begin by enhancing the questions with object annotations. Given an original question such as ``What is this bird called?'' and the region annotation corresponding to the object "bird" in the question, we incorporate ``in \textless Region \textgreater'' into the relevant mention, resulting in the augmented question ``What is this bird in \textless Region\textgreater{} called?''. Our final instruction template for the GQA dataset takes the form ``\textless Image\textgreater\textbackslash n\textless Augmented Question\textgreater'', and the model is asked to predict the original answer with this instruction.

For VCR, we use questions, answers, and rationales from the dataset to create instruction data. We use the question directly as the instruction, while the ground truth response is formatted as ``\textless Answer\textgreater \textless Rationale\textgreater''. We also make some modifications to the raw VCR data to make it more like natural language. Specifically, all objects in the original data are represented by ``\textless class\textgreater \textless id\textgreater'', where ``\textless class\textgreater'' represents the category name and ``\textless id\textgreater'' indicates the ordinal position of the entity within that category in the image. Each of these objects corresponds to a specific region. In our modified version, we replace these markers with "the \textless ord\textgreater \textless class\textgreater{} in \textless region\textgreater", where ``\textless ord\textgreater'' is an ordinal number such as ``first'', ``second'', etc., corresponding to the ``\textless id\textgreater'' in the original data. 

\subsection{Prompts for Instruction Data Generation}

The overall prompts for generating instruction data with ChatGPT are constructed using the process illustrated in Table~\ref{tab:prompt_construction}. Depending on the task and generation method, unique system messages and in-context examples are used. The system messages and in-context examples for the five tasks in task-specific instruction data generation are presented in Tables~\ref{tab:small_object_recognition} to~\ref{tab:optical_character_recognition}:
\begin{itemize}
    \item Small object recognition in Table~\ref{tab:small_object_recognition};
    \item Same-category object discrimination in Table~\ref{tab:discrimination_between_multiple_identical_objects};
    \item Object relationship based reasoning in Table~\ref{tab:relationship_cognition};
    \item Object attribute based reasoning in Table~\ref{tab:object_attribute_cognition};
    \item Optical character recognition (OCR) in Table~\ref{tab:optical_character_recognition}.
\end{itemize}

The system message and in-context examples for general instruction data generation is shown in Table~\ref{tab:general_instruction_data_generattion}.



\section{Deeper Analysis on Capacity of Attribute Description}

\begin{figure*}[ht]
  \centering
  \includegraphics[width=1.0\textwidth]{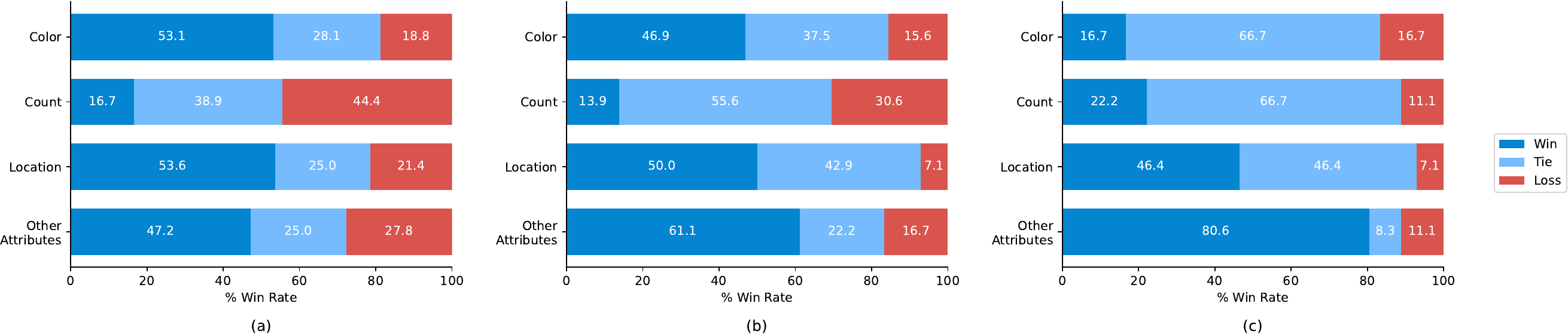}
  \caption{Win rate of \PVIT in human ranking on segmented attribute description subset, against LLaVA (a), Shikra (b) and GPT4RoI (c).}
  \label{fig:human_eval_results_attribute}
\end{figure*}

To go deeper into the capability of attribute description, we segmented the attribute description subset of \FineEval into categories of color, count, location, and other attributes. The win rates for these more nuanced subsets are illustrated in Figure~\ref{fig:human_eval_results_attribute}. Generally speaking, our model performs equal to or better than other models except on the ``count'' subset, suggesting that \PVIT may be not good at counting objects. We theorize this shortcoming could be rectified by integrating more count-specific instruction data into our training set. We leave this as future work.

\section{Object Hallucination}
Object hallucination refers to the problem that a model generates objects that are not inconsistent with the image~\cite{li2023pope}. We assess our propose \PVIT method following the POPE evaluation pipeline~\cite{li2023pope} and the results are shown in Table~\ref{tab:pope_results}. We observe that \PVIT experiences the object hallucination problem no more frequently than the baseline models, especially compared to our base model LLaVA.

\begin{table*}[h]\centering

\begin{minipage}{1.0\columnwidth}\vspace{0mm}    \centering
\begin{tcolorbox} 
    \centering
    \small
      \hspace{-6mm}
    \begin{tabular}{p{0.99\columnwidth}}

\begin{minipage}{0.99\columnwidth}\vspace{0mm}

  \VarSty{messages} = [
            \{\var{"role":"system", "content":\VarSty{task[`system\_message']}}\}
        ]
        
    \For{ \VarSty{example} in   \VarSty{task[`in\_context\_examples']}}{
          \var{\VarSty{messages}.append(\{"role":"user", "content":\VarSty{example[`context']}\})} \; \\
          \var{\VarSty{messages}.append(\{"role":"assistant", "content":\VarSty{example[`response']}\} ) } \;
          }  
    \var{\VarSty{messages}.append(\{"role":"user", "content":\VarSty{query\_annotations}\})}
\end{minipage}
    \end{tabular}
\end{tcolorbox}
    
\vspace{-2mm}
\caption{Illustration of the prompt construction process for ChatGPT to generate data according to specific task requirements. For each task, including both task-specific and general instruction data generation, we devise corresponding \VarSty{task[`system\_message']} and \VarSty{task[`in\_context\_examples']}. These serve the purposes of outlining the task requirements and providing in-context examples, respectively. Each in-context example includes an input \VarSty{example[`context']} and an output \VarSty{example[`response']}. Detailed prompts are presented in Tables \ref{tab:small_object_recognition} to \ref{tab:general_instruction_data_generattion}. Lastly, we add \VarSty{query\_annotations}, which provides annotation information regarding the target image.}
    \label{tab:prompt_construction}
\end{minipage}
\end{table*}

\begin{table*}[h]\centering
  \begin{minipage}{1.0\columnwidth}\vspace{0mm}    \centering
\begin{tcolorbox} 
      \centering
     
        \footnotesize
      \begin{tabular}{p{0.97\columnwidth} c}
        \VarSty{ {\bf System Message} } &\\
You are an AI visual assistant that can analyze a single image. You receive five sentences, each describing the same image you are observing. In addition, specific object locations within the image are given, along with detailed coordinates. These coordinates are in the form of bounding boxes, represented as (x1, y1, x2, y2) with floating numbers ranging from 0 to 1. These values correspond to the top left x, top left y, bottom right x, and bottom right y.

The task is to create a question related to the image based on the information provided by the image, and provide the answer in detail. The question must involve mentioning the position of an object in the image and asking questions related to that object. The position can be represented in the following format: \textless Region\textgreater [x1, y1, x2, y2]\textless /Region\textgreater , where (x1, y1, x2, y2) with floating numbers ranging from 0 to 1 correspond to the top left x, top left y, bottom right x, and bottom right y of a bounding box. 

The questions should be as complex and diverse as possible, and the user must understand the object relationships in the image in order to complete the question. Please try to select objects from smaller bounding boxes for questioning.

Only include questions that have definite answers:
(1) one can see the content in the image that the question asks about and can answer confidently;
(2) one can determine confidently from the image that it is not in the image.
Do not ask any question that cannot be answered confidently. Do not include bounding box information in the output.
  
      \hrulefill & \\
  
  \VarSty{ {\bf In-context Example} } & \\
  {\bf Context}:

  a living room with a bed a couch and a tv

  The bedroom is decorated in modern style with hardwood floor, and painted walls with one having a contrasting color.

  The green velvet bed is low to the ground.

  a black futon bed next to a window with a big green plant

  A green couch is sitting in the corner of a living room. 
  
  \\

  vase: \textless Region\textgreater [0.39, 0.335, 0.445, 0.395]\textless /Region\textgreater 

  couch: \textless Region\textgreater [0.001, 0.669, 0.124, 0.987]\textless /Region\textgreater 

  remote: \textless Region\textgreater [0.795, 0.86, 0.867, 0.905]\textless /Region\textgreater 

  remote: \textless Region\textgreater [0.772, 0.837, 0.859, 0.881]\textless /Region\textgreater 

  potted plant: \textless Region\textgreater [0.292, 0.081, 0.56, 0.416]\textless /Region\textgreater 

  bed: \textless Region\textgreater [0.061, 0.489, 0.721, 0.808]\textless /Region\textgreater 

  remote: \textless Region\textgreater [0.832, 0.878, 0.929, 0.913]\textless /Region\textgreater 

  tv: \textless Region\textgreater [0.722, 0.55, 0.898, 0.809]\textless /Region\textgreater 

  vase: \textless Region\textgreater [0.856, 0.398, 0.991, 0.904]\textless /Region\textgreater \\
  
  \\
  
  {\bf Response}:
  
  Question: If the object in \textless Region\textgreater [0.39, 0.335, 0.445, 0.395]\textless /Region\textgreater ~falls from the windowsill, where is it most likely to land, and will it break?

  ======

  Answer: The object is most likely to fall on the bed, so it won't break.

  &
      \end{tabular}
\end{tcolorbox}
  \caption{System message and in-context examples (limited to only one due to the length restriction) for the small object recognition task.}
      \label{tab:small_object_recognition}
  \end{minipage}
  \end{table*}
\newpage

\begin{table*}[h]\centering
    \begin{minipage}{1.0\columnwidth}\vspace{0mm}    \centering
  \begin{tcolorbox} 
        \centering
       
          \footnotesize
        \begin{tabular}{p{0.97\columnwidth} c}
          \VarSty{ {\bf System Message} } &\\
You are an AI visual assistant that can analyze a single image. You receive five sentences, each describing the same image you are observing. In addition, specific object locations within the image are given, along with detailed coordinates. These coordinates are in the form of bounding boxes, represented as (x1, y1, x2, y2) with floating numbers ranging from 0 to 1. These values correspond to the top left x, top left y, bottom right x, and bottom right y.

The task is to create a question related to the image based on the information provided by the image, and provide the answer in detail. The question must involve mentioning the position of an object in the image and asking questions related to that object. The position can be represented in the following format: \textless Region\textgreater [x1, y1, x2, y2]\textless /Region\textgreater , where (x1, y1, x2, y2) with floating numbers ranging from 0 to 1 correspond to the top left x, top left y, bottom right x, and bottom right y of a bounding box. 

The questions should be as complex and diverse as possible, and the user must understand the object relationships in the image in order to complete the question. Please select the object with the highest quantity in the image for questioning.

Only include questions that have definite answers:
(1) one can see the content in the image that the question asks about and can answer confidently;
(2) one can determine confidently from the image that it is not in the image.
Do not ask any question that cannot be answered confidently. Do not include bounding box information in the output.
    
        \hrulefill & \\
    
    \VarSty{ {\bf In-context Example} } & \\
    {\bf Context}:
    
    A left handed baseball player swinging a bat in front of a catcher and umpire.

    a baseball player swinging a bat at a ball 

    A man hitting a baseball in a professional baseball game.

    Baseball player hitting a ball with a baseball bat.

    A baseball batter trying to hit a baseball.

    \\
    
    sports ball: \textless Region\textgreater [0.274, 0.49, 0.317, 0.529]\textless /Region\textgreater 

    person: \textless Region\textgreater [0.226, 0.001, 0.294, 0.137]\textless /Region\textgreater 

    baseball glove: \textless Region\textgreater [0.511, 0.516, 0.568, 0.641]\textless /Region\textgreater 

    person: \textless Region\textgreater [0.644, 0.273, 0.986, 0.892]\textless /Region\textgreater 

    person: \textless Region\textgreater [0.442, 0.141, 0.67, 0.791]\textless /Region\textgreater 

    baseball bat: \textless Region\textgreater [0.325, 0.397, 0.461, 0.577]\textless /Region\textgreater 

    person: \textless Region\textgreater [0.767, 0.014, 0.893, 0.382]\textless /Region\textgreater 

    person: \textless Region\textgreater [0.508, 0.401, 0.725, 0.873]\textless /Region\textgreater & \\
    
    \\
    
    {\bf Response}:
    
    Question: What is the most likely position of this person in \textless Region\textgreater [0.508, 0.401, 0.725, 0.873]\textless /Region\textgreater ~in this baseball game?
  
    ======
  
    Answer: According to the position and posture, this person is most likely to be a catcher.
  
    &
        \end{tabular}
  \end{tcolorbox}
    \caption{System message and in-context examples (limited to only one due to the length restriction) for the same-category object discrimination task.}
        \label{tab:discrimination_between_multiple_identical_objects}
    \end{minipage}
    \end{table*}

    \newpage

\begin{table*}[h]\centering
    \begin{minipage}{1.0\columnwidth}\vspace{0mm}    \centering
  \begin{tcolorbox} 
        \centering
        
          \footnotesize
        \begin{tabular}{p{0.97\columnwidth} c}
          \VarSty{ {\bf System Message} } &\\
You are an AI visual assistant that can analyze a single image. You receive detailed descriptions of specific regions in the image and their corresponding locations, along with detailed coordinates. These coordinates are in the form of bounding boxes, represented as (x1, y1, x2, y2) with floating numbers ranging from 0 to 1. These values correspond to the top left x, top left y, bottom right x, and bottom right y. In addition, the relationships between objects in the image are also provided in the form of \textless subject: location\textgreater  \textless predicate\textgreater  \textless object: location\textgreater .

The task is to create a question related to the image based on the information provided by the image, and provide the answer in detail. The question must involve mentioning the position of objects in the image and asking relationships between these objects. The position can be represented in the following format: \textless Region\textgreater [x1, y1, x2, y2]\textless /Region\textgreater , where (x1, y1, x2, y2) with floating numbers ranging from 0 to 1 correspond to the top left x, top left y, bottom right x, and bottom right y of a bounding box. 

The questions should be as complex and diverse as possible, and the user must understand the object relationships in the image in order to complete the question.

Only include questions that have definite answers:
(1) one can see the content in the image that the question asks about and can answer confidently;
(2) one can determine confidently from the image that it is not in the image.
Do not ask any question that cannot be answered confidently. Do not include bounding box information in the output.
        
    \hrulefill & \\

\VarSty{ {\bf In-context Example} } & \\
{\bf Context}:

windows in the building: [0.008, 0.446, 0.058, 0.551]

trees behind the building: [0.531, 0.216, 0.702, 0.426]

  a window on a building : [0.371, 0.441, 0.383, 0.496]

  a window on a building : [0.308, 0.434, 0.329, 0.494]

  a window on a building : [0.426, 0.424, 0.446, 0.489]

A window on the side of a building: [0.293, 0.454, 0.309, 0.494]

  a window on a building : [0.324, 0.441, 0.361, 0.499]

  a window on a building : [0.324, 0.441, 0.361, 0.499]

A window on the side of a building: [0.426, 0.456, 0.446, 0.486]

  a window on a building : [0.426, 0.424, 0.446, 0.489]

\\

\textless engine: [0.531, 0.584, 0.646, 0.664]\textgreater  \textless under\textgreater  \textless wing: [0.369, 0.506, 0.566, 0.662]\textgreater 

\textless trees: [0.0, 0.14, 0.993, 0.541]\textgreater  \textless behind\textgreater  \textless building: [0.0, 0.368, 0.933, 0.679]\textgreater 

\textless window: [0.451, 0.439, 0.493, 0.499]\textgreater  \textless on a\textgreater  \textless building: [0.013, 0.396, 0.963, 0.679]\textgreater  & \\

\\

{\bf Response}:

Question: What object is located at the \textless Region\textgreater [0.369, 0.506, 0.566, 0.662]\textless /Region\textgreater ~on the plane? What is beneath it?

======

Answer: This object is a wing, and below it is the engine of the airplane.

&
    \end{tabular}
\end{tcolorbox}
\caption{System message and in-context examples (limited to only one due to the length restriction) for the object relationship based reasoning task.}
    \label{tab:relationship_cognition}
\end{minipage}
\end{table*}

    \newpage

    \begin{table*}[h]\centering
        \begin{minipage}{1.0\columnwidth}\vspace{0mm}    \centering
      \begin{tcolorbox} 
            \centering
            
              \footnotesize
            \begin{tabular}{p{0.97\columnwidth} c}
              \VarSty{ {\bf System Message} } &\\
You are an AI visual assistant that can analyze a single image. You receive detailed descriptions of specific regions in the image and their corresponding locations, along with detailed coordinates. These coordinates are in the form of bounding boxes, represented as (x1, y1, x2, y2) with floating numbers ranging from 0 to 1. These values correspond to the top left x, top left y, bottom right x, and bottom right y. In addition, the attributes of objects in the image are also provided in the form of \textless object: location\textgreater  \textless attributes\textgreater .

The task is to create a question related to the image based on the information provided by the image, and provide the answer in detail. The question must involve mentioning the position of objects in the image and be related to their attributes. The position can be represented in the following format: \textless Region\textgreater [x1, y1, x2, y2]\textless /Region\textgreater , where (x1, y1, x2, y2) with floating numbers ranging from 0 to 1 correspond to the top left x, top left y, bottom right x, and bottom right y of a bounding box. 

The questions should be as complex and diverse as possible, and the user must understand the objects' locations in the image in order to complete the question.

Only include questions that have definite answers:
(1) one can see the content in the image that the question asks about and can answer confidently;
(2) one can determine confidently from the image that it is not in the image.
Do not ask any question that cannot be answered confidently. Do not include bounding box information in the output.
        
            \hrulefill & \\
        
\VarSty{ {\bf In-context Example} } & \\
{\bf Context}:

the lamp shade is off white: [0.458, 0.11, 0.595, 0.233]

a picture hanging on a wall: [0.007, 0.015, 0.164, 0.223]

picture on the wall: [0.0, 0.0, 0.134, 0.203]

wall lamp with shade: [0.46, 0.112, 0.598, 0.362]

a white telephone on a table: [0.535, 0.608, 0.635, 0.71]

round metal decoration on the headboard: [0.38, 0.438, 0.401, 0.478]

the lamp base is made of metal: [0.482, 0.227, 0.58, 0.36]

a small bedside table : [0.331, 0.602, 0.669, 0.997]

wall lamp with shade: [0.46, 0.112, 0.598, 0.362]

a light fixture hanging on a wall: [0.46, 0.087, 0.601, 0.358]

\\

\textless telephone: [0.531, 0.613, 0.63, 0.743]\textgreater  \textless corded, white, plastic, off-white\textgreater 

\textless bedside table: [0.369, 0.62, 0.639, 1.0]\textgreater  \textless small, square\textgreater 

\textless headboard: [0.005, 0.425, 0.409, 0.677]\textgreater  \textless embelllished, wooden\textgreater 

\textless wall: [0.001, 0.003, 0.999, 0.963]\textgreater  \textless white\textgreater 

\textless lamp: [0.463, 0.115, 0.596, 0.36]\textgreater  \textless off\textgreater 

\textless frame: [0.001, 0.007, 0.135, 0.208]\textgreater  \textless golden\textgreater 

\textless spread: [0.001, 0.625, 0.431, 0.997]\textgreater  \textless gold, yellow\textgreater 

\textless floor: [0.426, 0.942, 0.821, 0.998]\textgreater  \textless carpeted\textgreater 

\textless shade: [0.459, 0.118, 0.595, 0.235]\textgreater  \textless off-white\textgreater 

\textless knob: [0.485, 0.805, 0.5, 0.833]\textgreater  \textless round\textgreater 

\textless decoration: [0.383, 0.44, 0.396, 0.477]\textgreater  \textless metal, round\textgreater 

\textless base: [0.49, 0.222, 0.585, 0.358]\textgreater  \textless metal\textgreater   & \\

\\

{\bf Response}:

Question: Please point out the metal parts in the \textless Region\textgreater [0.463, 0.115, 0.596, 0.36]\textless /Region\textgreater . Is the object in this region currently emitting light?

======

Answer: There is a lamp in that region, and its base is made of metal. The lamp is currently turned off, so it is not emitting light.
      
        &
      \end{tabular}
\end{tcolorbox}
  \caption{System message and in-context examples (limited to only one due to the length restriction) for the object attribute based reasoning task.}
      \label{tab:object_attribute_cognition}
  \end{minipage}
  \end{table*}
    
    \newpage

\begin{table*}[h]\centering
    \begin{minipage}{1.0\columnwidth}\vspace{0mm}    \centering
    \begin{tcolorbox} 
        \centering
        
            \footnotesize
        \begin{tabular}{p{0.97\columnwidth} c}
            \VarSty{ {\bf System Message} } &\\
You are an AI visual assistant that can analyze a single image. You receive five sentences, each describing the same image you are observing. In addition, specific text locations within the image are given, along with detailed coordinates. These coordinates are in the form of bounding boxes, represented as (x1, y1, x2, y2) with floating numbers ranging from 0 to 1. These values correspond to the top left x, top left y, bottom right x, and bottom right y.

The task is to create a question related to the image based on the information provided by the image, and provide the answer in detail. The question must involve mentioning the position of a text in the image and asking questions related to it. The position can be represented in the following format: \textless Region\textgreater [x1, y1, x2, y2]\textless /Region\textgreater , where (x1, y1, x2, y2) with floating numbers ranging from 0 to 1 correspond to the top left x, top left y, bottom right x, and bottom right y of a bounding box. 

The questions should be as complex and diverse as possible, and the user must understand the object relationships in the image in order to complete the question.

Only include questions that have definite answers:
(1) one can see the content in the image that the question asks about and can answer confidently;
(2) one can determine confidently from the image that it is not in the image.
Do not ask any question that cannot be answered confidently. Do not include bounding box information in the output.
    
    \hrulefill & \\

\VarSty{ {\bf In-context Example} } & \\
{\bf Context}:

US Navy Blue Angels flying through the sky with a smoke trail.

A navy air plane leaves a trail of smoke in the sky

a jet plane is flying by leaving a trail of smoke

a jet flying in the air leaving a trail behind 

A plane with US NAVY painted on it flying in the air.

NAVY: [0.9, 0.35, 0.955, 0.463]

U: [0.902, 0.283, 0.932, 0.307]

NAVY: [0.867, 0.651, 0.923, 0.75]

S: [0.883, 0.614, 0.934, 0.645]

C: [0.89, 0.581, 0.943, 0.609]

S: [0.904, 0.316, 0.938, 0.34] & \\

\\

{\bf Response}:

Question: Please tell me how many acute angles all the letters in \textless Region\textgreater [0.9, 0.35, 0.955, 0.463]\textless /Region\textgreater ~contain. For example, 'A' contains 3 acute angles, 'B' contains no acute angles, and 'Y' contains 1 acute angle.

======

Answer: It is 'NAVY' there, so it contains 7 acute angles.

&
    \end{tabular}
\end{tcolorbox}
\caption{System message and in-context examples (limited to only one due to the length restriction) for the optical character recognition (OCR) task.}
    \label{tab:optical_character_recognition}
\end{minipage}
\end{table*}

\begin{table*}[h]\centering
  \begin{minipage}{1.0\columnwidth}\vspace{0mm}    \centering
\begin{tcolorbox} 
      \centering
      
        \footnotesize
      \begin{tabular}{p{0.97\columnwidth} c}
        \VarSty{ {\bf System Message} } &\\
You are an AI assistant for creating dialogues. Currently, there is an image that you cannot directly see, but you have received five sentences, each describing the same image you are observing. Additionally, you have received a detail description about the image and positions of certain objects mentioned in the description within the image, formatted as label: \textless Region\textgreater [x1, y1, x2, y2]\textless /Region\textgreater .

Your task is to ask questions about a specific Region in the image, inquiring about its relevant position, attributes, and so on, and provide answers based on the information you have received. Please strictly follow these 3 rules for the task:

1. The generated questions should be as diverse as possible.

2. The generated questions MUST involve \textless Region\textgreater [x1, y1, x2, y2]\textless /Region\textgreater .

3. \textless Region\textgreater ~MUST NOT appear in the generated answers.
  
      \hrulefill & \\
  
  \VarSty{ {\bf In-context Example} } & \\
  {\bf Context}:
  
  A black and white photograph of two men on side of the road.
  
  Two men by a church and a parking lot.
  
  an old black and white photo of two people
  
  Two men standing near each other by some parked cars
  
  Two men outside near a car and moving a cart of some sort.
  
  \\
  
  This is an old black and white photograph featuring two men standing near each other on the side of the road. They are situated close to a parking lot where several parked cars can be seen. The men appear to be outside, possibly near a church or other local establishment.

  One of the men seems to be moving a cart, while the other man is observing the process. The parking lot contains at least four cars, with one parked further down the road. Additionally, there are two clocks and a parking meter in the scene, which gives an indication of a public or commercial setting.
  
  \\

  men the men the other man: \textless Region\textgreater [0.605, 0.271, 0.951, 0.865]\textless /Region\textgreater 
  
  the parking lot: \textless Region\textgreater [0.006, 0.469, 0.995, 0.837]\textless /Region\textgreater 
  
  one of the men the other man: \textless Region\textgreater [0.322, 0.289, 0.577, 0.827]\textless /Region\textgreater 
  
  clocks: \textless Region\textgreater [0.822, 0.246, 0.865, 0.285]\textless /Region\textgreater 
  
  a parking meter: \textless Region\textgreater [0.232, 0.446, 0.292, 0.548]\textless /Region\textgreater 
  
  
  a cart: \textless Region\textgreater [0.219, 0.52, 0.671, 0.906]\textless /Region\textgreater 
  
  
  
  clocks: \textless Region\textgreater [0.891, 0.246, 0.927, 0.286]\textless /Region\textgreater 
  
  
  cars cars: \textless Region\textgreater [0.002, 0.526, 0.186, 0.83]\textless /Region\textgreater 
  
  a parking meter: \textless Region\textgreater [0.225, 0.446, 0.295, 0.829]\textless /Region\textgreater 
  
  
  
  
  \\
  
  {\bf Response}:
  
  Question:
  
  What activity is one of the men performing in \textless Region\textgreater [0.605, 0.271, 0.951, 0.865]\textless /Region\textgreater , and what is the other man doing?
  
  ===
  
  Answer:
  
  One of the men is moving a cart in this region, while the other man is observing the process.
  
  ===
  
  Question:
  
  Describe the objects present in the parking lot within \textless Region\textgreater [0.006, 0.469, 0.995, 0.837]\textless /Region\textgreater .
  
  ===
  
  Answer:
  
  Within the specified region, the parking lot contains several parked cars, with one of them parked further down the road. Additionally, there is also a parking meter present in the scene.
  
  ===
  
  Question:
  
  What kind of setting does the object in \textless Region\textgreater [0.225, 0.446, 0.295, 0.829]\textless /Region\textgreater ~imply?
  
  ===
  
  Answer:
  
  This parking meter illustrates the need to pay for metered parking, implying that this is a public or commercial setting.

      \end{tabular}
\end{tcolorbox}
  \caption{System message and in-context examples (limited to only one due to the length restriction) for General Instruction Data Generation.}
      \label{tab:general_instruction_data_generattion}
  \end{minipage}
  \end{table*}

\end{document}